\documentclass[runningheads]{llncs}
\usepackage{graphicx}
\usepackage{amsfonts}
\usepackage{amsmath}
\usepackage{amssymb}
\usepackage{bm}
\usepackage{marvosym}
\usepackage{indentfirst}
\usepackage{multirow, makecell}
\usepackage{booktabs}
\usepackage{tabu}
\usepackage[normalem]{ulem}
\usepackage[table,xcdraw]{xcolor}
\usepackage[section]{placeins}
\usepackage{rotfloat}
\usepackage[colorlinks,linkcolor=blue,citecolor=blue]{hyperref}

\begin{document}
\title{UnDeepLIO: Unsupervised Deep Lidar-Inertial Odometry}
%
%
\author{Yiming Tu\inst{1,2} \and
Jin Xie \textsuperscript{1,2(\Letter)}}
\authorrunning{Y. Tu \emph{et al.}}
%
\institute{PCA Lab, Key Lab of Intelligent Perception and Systems for High-Dimensional
Information of Ministry of Education, Nanjing University of Science and Technology,\\
Nanjing, China\\
\email{\{tymstudy,csjxie\}@njust.edu.cn}
\and
Jiangsu Key Lab of Image and Video Understanding for Social Security,\\
School of Computer Science and Engineering,\\
Nanjing University of Science and Technology, Nanjing, China}
\maketitle              

\begin{abstract}
Extensive research efforts have been dedicated to deep learning based odometry. 
Nonetheless, few efforts are made on the unsupervised deep lidar odometry. 
In this paper, we design a novel framework for unsupervised lidar odometry with 
the IMU, which is never used in other deep methods. First, a pair of siamese 
LSTMs are used to obtain the initial pose from the linear acceleration and 
angular velocity of IMU. With the initial pose, we perform the rigid transform 
on the current frame and align it to the last frame. Then we extract 
vertex and normal features from the transformed point clouds and its normals. 
Next a two-branch attention module is proposed to estimate residual rotation 
and translation from the extracted vertex and normal features, respectively. 
Finally, our model outputs the sum of initial and residual poses as the final 
pose. For unsupervised training, we introduce an unsupervised loss function 
which is employed on the voxelized point clouds. The proposed approach is 
evaluated on the KITTI odometry estimation benchmark and achieves comparable 
performances against other state-of-the-art methods.
\end{abstract}

\section{Introduction}
The task of odometry is to estimate 3D translation and orientation 
of autonomous vehicles which is one of key steps in SLAM. 
Autonomous vehicles usually 
collect information by perceiving the surrounding environment in real time and 
use on-board sensors such as lidar, Inertial Measurement Units (IMU), or 
camera to estimate their 3D translation and orientation. 
Lidar can provide high-precision 3D measurements but also has no 
requirement for light. 
The point clouds generated by the lidar can provide high-precision 3D measurements, 
but if it has large translation or orientation in a short time, the 
continuously generated point clouds will only get few matching points, which 
will affect the accuracy of odometry. 
IMU has advantages of high output frequency and 
directly outputting the 6DOF information to predict the initial translation 
and orientation that the 
localization failure phenomenon can be reduced when lidar has large 
translation or orientation.\\
\indent The traditional methods \cite{ICP,GICP,Loam,Lego-loam} 
are mainly based on the point registration and work well in ideal environments. 
However, due to the sparseness and irregularity of the point clouds, these 
methods are difficult to obtain enough matching points. Typically, ICP 
\cite{ICP} and its variants \cite{GICP,po2plICP} iteratively find 
matching points which depend on nearest-neighbor searching and optimize the 
translation and orientation by matching points. This optimization procedure is 
sensitive to noise and dynamic objects and prone to getting stuck into the 
local minima.\\
\indent Thanks to the recent advances in deep learning, many 
approaches adopt deep neural networks for lidar odometry, which can achieve 
more promising performance compared to traditional methods. 
But most of them are supervised methods 
\cite{Lo-net,LodoNet,DeepPCO,DMLO,Velas,DeepLIO}. 
However, supervised methods require ground truth pose, 
which consumes a lot of manpower and material resources. 
Due to the scarcity of the ground truth, recent unsupervised methods are proposed
\cite{Cho,Nubert,SelfVoxeLO}, 
but some of them obtain unsatisfactory performance, and some need to consume a lot of 
video memory and time to train the network.\\
\indent Two issues exist in these methods. First, these methods ignore 
IMU, which often bring fruitful clues for accurate lidar odometry.  
Second, those methods do not make full use of the normals, 
which only take the point clouds as the inputs. Normals of point clouds can indicate the 
relationship between a point and its surrounding points. And even if 
those approaches \cite{Lo-net} who use normals as network input, they 
simply concatenate points and normals together and put them into 
network, but only orientation between two point clouds relates to normals, 
so normals should not be used to estimate translation.\\
\indent To circumvent the dependence on expensive ground truth, 
we propose a novel framework termed as UnDeepLIO, which makes full use of 
the IMU and normals for more accurate odometry. We compare against various baselines 
using point clouds from the KITTI Vision Benchmark Suite \cite{KITTI} which 
collects point clouds using a $360^\circ$ Velodyne laser scanner.\\
\indent Our main contributions are as follows:\\
\indent $\bullet $ We present a self-supervised learning-based approach for 
robot pose estimation. our method can outperform \cite{Cho,Nubert}.\\
\indent $\bullet $ We use IMU to assist odometry. 
Our IMU feature extraction module can be embedded in most network structures 
\cite{Lo-net,DeepPCO,Cho,Nubert}. \\
\indent $\bullet $ Both points and its normals are used as network inputs. 
We use feature of points to estimate translation and 
feature of both of them to estimate orientation. \\
\section{Related Work}
\subsection{Model-based Odometry Estimation}
Gauss-Newton iteration methods have a long-standing history in odometry task. 
Model-based methods solve odometry problems generally by using Newton's 
iteration method to adjust the transformation between frames so that 
the "gap" between frames keeps getting smaller. They can be categorized into 
two-frame methods \cite{ICP,GICP,po2plICP} and multi-frame methods 
\cite{Loam,Lego-loam}.\\
\indent Point registration is the most common skill for two-frame methods, where 
ICP \cite{ICP} and its variants \cite{GICP,po2plICP} are typical 
examples. The ICP iteratively search key points and its correspondences 
to estimate the transformation between two point clouds until 
convergence. Moreover, most of 
these methods need multiple iterations with a large amount of calculation, 
which is difficult to meet the real-time requirements of the system.\\
\indent Multi-frame algorithms \cite{Loam,Lego-loam,SUMA} often relies 
on the two-frame based estimation. They improve the steps of selecting key 
points and finding matching points, and use additional mapping step to 
further optimize the pose estimation. Their calculation process is 
generally more complicated and runs at a lower frequency.

\subsection{Learning-based Odometry Estimation}
In the last few years, the development of deep learning has greatly affected 
the most advanced odometry estimation. Learning-based model can provide a 
solution only needs uniformly down sampling the point clouds without manually selecting 
key points. They can be classified into supervised 
methods and unsupervised methods.\\
\indent Supervised methods appear relatively early, Lo-net \cite{Lo-net} maps the point 
clouds to 2D "image" by spherical projection. Wang \emph{et al.} \cite{DeepPCO} adopt a dual-branch 
architecture to infer 3-D translation and orientation separately instead of a single network. Velas 
\emph{et al.} \cite{Velas} use point clouds to assist 3D motion estimation and 
regarded it as a classification problem. Differently, Li \emph{et al.} \cite{DMLO} 
do not simply estimate 3D motion with fully connected layer but Singular 
Value Decomposition (SVD). Use Pointnet \cite{PointNet} as base net, 
Zheng \emph{et al.} \cite{LodoNet} propose a new approach for extracting matching 
keypoint pairs(MKPs) of consecutive LiDAR scans by projecting 3D point clouds 
into 2D spherical depth images where MKPs can be extracted effectively and 
efficiently.\\ 
\indent Unsupervised methods appear later. Cho \emph{et al.} \cite{Cho} first 
apply unsupervised approach on deep-learning-based LiDAR odometry which is 
an extension of their previous approach \cite{DeepLo}. The inspiration of its 
loss function comes from point-to-plane ICP \cite{po2plICP}. Then, Nubert \emph{et al.} 
\cite{Nubert} report methods with similarly models and loss function, 
but they use different way to calculate normals of each point in point clouds 
and find matching points between two continuous point clouds.

\section{Methods}
\subsection{Data Preprocess}
\subsubsection{Data input}
At every timestamp $k\in\mathbb{R}^+$, we can obtain one point clouds $P_k$ of 
$N*3$ dimensions and between every two timestamps we can get $S$ frames IMU 
$I_{k,k+1}$ of $S*6$ dimensions including 3D angular velocity and 3D linear acceleration 
. we take above data as the inputs.

\subsubsection{Vertex map}
In order to circumvent the disordered nature of point clouds, 
we project the point clouds into the 2D image coordinate system 
according to the horizontal and vertical angle. We employ 
projection function $\Pi:{\mathbb{R}}^3\mapsto{\mathbb{R}}^2$. Each 3D point 
$\bm{p}=(p_x,p_y,p_z)$ in a point clouds $P_k$ is mapped into the 2D image plane 
$(w,h)$ represented as

\begin{equation}
\begin{split}       
    \left(                
        \begin{array}{c}   
            w \\
            h \\
        \end{array}
    \right)                 
    =&
    \left(                 
        \begin{array}{c}   
            (f_w-\arctan(\frac{p_y}{p_x}))/\eta_w \\  
            (f_h-\arcsin(\frac{p_z}{d}))/\eta_h\\  
        \end{array}
    \right),                 \\
    & H > h \geq 0, W > w \geq 0,
\end{split}
\end{equation}

where depth is $d=\sqrt{{p_x}^2+{p_y}^2+{p_z}^2}$. $f_w$ and $f_h$ 
are the maximum horizontal and vertical angle. $H$ and $W$ are shape of vertex map. $f_h$ depends 
on the type of the lidar. $\eta_w$ and $\eta_h$ 
control the horizontal and vertical sampling density. If several 3D points 
correspond the same pixel values, we choose the point with minimum depth as the final result. 
If one pixel coordinate has no matching 3D points, the 
pixel value is set to $(0,0,0)$. We define the 2D image plane as 
vertex map $\bm{V}$.

\subsubsection{Normal map}
The normal vector of one point includes its relevance about the surrounding points, 
so we compute a normal map $\bm{N}$ which consists of normals 
$\bm{n}$ and has the same shape as corresponding vertex map $\bm{V}$.
We adopt similar operations with Cho \emph{et al.} \cite{Cho} and Li \emph{et al.} \cite{Lo-net}
to calculate the normal vectors. 
Each normal vector $\bm{n}$ corresponds to a vertex $\bm{v}$ with the same image 
coordinate. Due to sparse and discontinuous characteristics of point clouds, 
we pay more attention on the vertex with small Euclidean distance from the 
surrounding pixel via a pre-defined weight, which can be expressed as 
$w_{a,b}=e^{\{-0.5|d(v_a)-d(v_b)|\}}$. Each normal vector 
$\bm{n}$ is represented as
\begin{equation}
\bm{n}_p=\sum_{i\in[0,3]}w_{p_i,p}(v_{p_i}-v_p)\times w_{p_{i+1},p}(v_{p_{i+1}}-v_p),
\end{equation}
where $p_i$ represents points in 4 directions 
of the central vertex $p$ (0-up, 1-right, 2-down, 3-left).

\subsection{Network Structure}

\begin{figure}[!t]
    \centering
    \includegraphics[width=\textwidth]{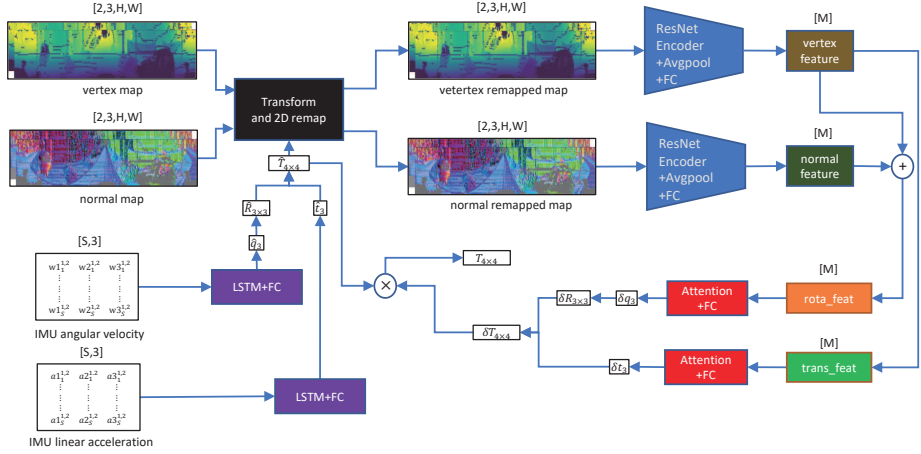}
    \caption{The proposed network and our unsupervised training scheme. 
    FC represents fully connected layer. $t$ means translation and $q$ means Euler angle
    of orientation. LSTM takes continuous frames of IMU as inputs and output 
    initial relative pose $\hat{T}$. $\hat{T}$ are used to transform two maps of 
    current frame to last frame. Then we send the remapped
    maps into ResNet Encoder, which outputs feature maps, including vertex and normal features. 
    From the features, we propose an attention layer to estimate residual pose $\delta T$. 
    The final output is their sum 
    $T=\delta T\hat{T}$.} 
    \label{fig1}
\end{figure}

\subsubsection{Network input}
Our pipeline is shown in the Fig.~\ref{fig1}. Each point clouds associates with a vertex/normal map of $(3,H,W)$ dimensions, 
so we concatenate the vertex/normal map of $k$ and $k+1$ timestamp to get vertex/normal pair 
of $(2,3,H,W)$ dimensions. We take a pair of vertex/normal
maps and IMU between $k$ and $k+1$ timestamp as the inputs of our model, where the IMU consists of the linear 
acceleration and angular velocity both of $(S,3)$ dimensions, and $S$ is the length of IMU sequence. 
Our model outputs relative pose $T_{k,k+1}$, where $R_{k,k+1}$ is orientation and $t_{k,k+1}$ 
is translation. 
\begin{equation}
T_{k,k+1}^{4\times4}=
\left[
\begin{array}{cc}
R_{k,k+1}^{3\times3} & t_{k,k+1}^{3\times1} \\
0 & 1
\end{array}
\right],
\end{equation}
\subsubsection{Estimating initial relative pose from IMU}
Linear acceleration is used to estimate translation and angular velocity is used to 
estimate orientation. We employ LSTM on IMU to extract the features of IMU. Then the 
features are forwarded into the FC layer to estimate initial relative translation or orientation.

\subsubsection{Mapping the point clouds of current frame to the last frame}
Each vertex/normal pair consists of last and current frames. They are not 
in the same coordinate due to the transformation. The initial relative pose can map current 
frame in current coordinate to last coordinate, then we can obtain the remapped current 
map with the same size as the old one. The relationship 
between two maps are shown as formula (\ref{formula1}). Take the $\bm{v}_{k+1,p}^{k}$ for example, it is the mapped 
vertex at timestamp $k$ from timestamp $k+1$ via the initial pose.
\begin{equation}
\bm{v}_{k+1,p}^{k}=R_{k,k+1}\bm{v}_{k+1,p}^{k+1}+t_{k,k+1}\label{formula1},
\end{equation}
\begin{equation}
\bm{n}_{k+1,p}^{k}=R_{k,k+1}\bm{n}_{k+1,p}^{k+1}.
\end{equation}

\subsubsection{Estimating residual relative pose from the remapped maps}
\begin{figure}[!t]
    \includegraphics[width=\textwidth]{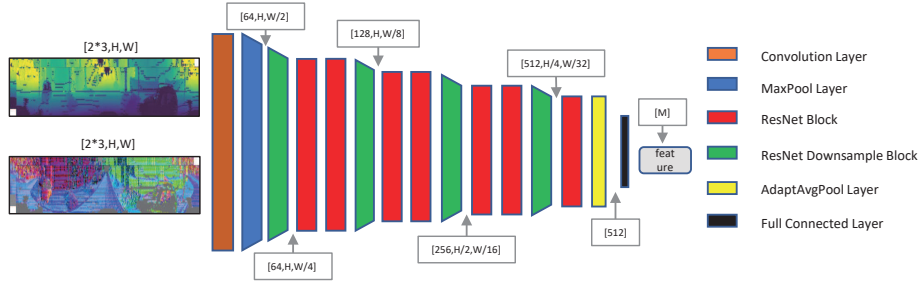}
    \caption{The detail structure of ResNet Encoder + Avgpool + FC part.} 
    \label{fig2}
\end{figure}
We use ResNet Encoder (see Fig.~\ref{fig2}) as our map feature extractor. 
ResNet \cite{ResNet} is used in image recognition. Its input is the 2D images similar to us. 
Therefore, this structure can extract feature in our task as well. 
we send the remapping vertex/normal map pair 
into siamese ResNet Encoder, which outputs feature maps, including vertex and normal 
features. From the features, we propose an attention layer 
(by formula (\ref{formula2}), $x$ is input) which is inspired by LSTM \cite{LSTM} 
to estimate residual pose $\delta T$ between last frame and the remapped current frame.
. Among them, vertex and 
normal features are combined to estimate orientation, but only vertex is used to estimate 
translation because the change of translation does not cause the change of normal vectors. 
Together with initial relative pose, we can get final relative pose $T$.
\begin{equation}
\begin{split}
&i=\sigma(W_{i}x+b_i), \\
&g=tanh(W_{g}x+b_g),  \\
&o=\sigma(W_{o}x+b_o),   \\
&out=o*tanh(i*g) \label{formula2}.
\end{split}
\end{equation}

\subsection{Loss Function}
\begin{figure}[!t]
    \includegraphics[width=\textwidth]{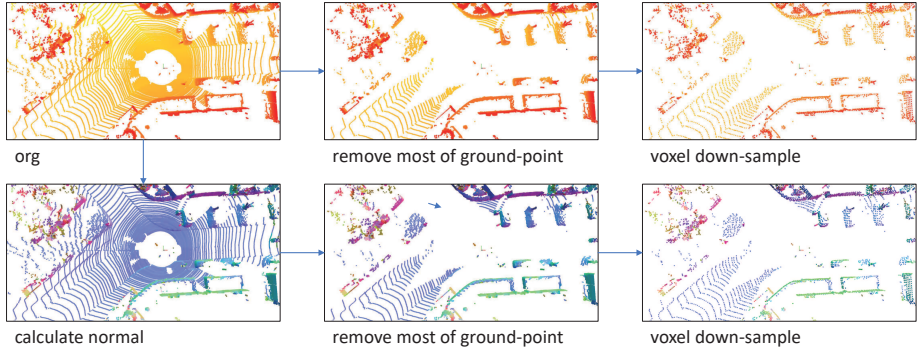}
    \caption{Point downsample process, including point (up) and normal (down).} 
    \label{fig0}
\end{figure}
For unsupervised training, we use a combination of geometric losses in our 
deep learning framework. Unlike Cho \emph{et al.} \cite{Cho} who use pixel locations as correspondence 
between two point clouds, we search correspondence on the whole point clouds. 
For speeding up calculation, we first calculate the normals $NP_i$ of whole 
point clouds $P_i$ by plane fitting $\Phi$ \cite{norestimate}, and then remove its ground points by RANSAC 
\cite{RANSAM}, at last perform voxel grid filtering $\Downarrow$ (the arithmetic average of all points in
voxel as its representation. The normal vectors of voxel are processed in the same way and standardized after downsample.) 
to downsample to about $K$ points (The precess is shown in Fig.~\ref{fig0}). Given the predicted relative pose $T_{k,k+1}$, we apply it on preprocessed 
current point clouds $DP_{k+1}$ and its normals $NP_{k+1}$. For the correspondence search, 
we use KD-Tree \cite{KDTree} to find the nearest point in the last point clouds $DP_k$ of each point 
in the transformed current point clouds $\overline{DP}_{k+1}$.
\begin{equation}
\begin{split}
&DP_{i}=\Downarrow(\mathsf{RANSAC}(P_{i})), \\
&NP_{i}=\Downarrow(\mathsf{RANSAC}(\Phi(P_{i})),
\end{split}
\end{equation}
\begin{equation}
\begin{split}
&\overline{NP}_{k+1}=R_{k,k+1}NP_{k+1}, \\ 
&\overline{DP}_{k+1}=R_{k,k+1}DP_{k+1}+t_{k,k+1}.
\end{split}
\end{equation}

\subsubsection{Point-to-plane ICP loss}
We use every point $\overline{dp}_{k+1}$ in current point clouds $\overline{DP}_{k+1}$, 
corresponding point of ${dp}_k$ and normal vector of ${np}_k$ in last point clouds ${DP}_{k}$ 
to compute the distance between point and its matching plane. The loss 
function $\mathcal{L}_{po2pl}$ is represented as
\begin{equation}
\mathcal{L}_{po2pl}=\sum_{\overline{dp}_{k+1}\in \overline{DP}_{k+1}} |{np}_k\centerdot(\overline{dp}_{k+1}-{dp}_k)|_1,
\end{equation}
where $\centerdot$ denotes element-wise product.
\subsubsection{Plane-to-plane ICP loss}
Similarly to point-to-plane ICP, we use normal $\overline{np}_{k+1}$ of every point in $\overline{NP}_{k+1}$, 
corresponding normal vector of ${np}_k$ in ${NP}_{k}$ to compute the angle between 
a pair of matching plane. The loss 
function $\mathcal{L}_{pl2pl}$ is represented as
\begin{equation}
\mathcal{L}_{pl2pl}=\sum_{\overline{np}_{k+1}\in \overline{NP}_{k+1}} |\overline{np}_{k+1}-{np}_k|^2_2.
\end{equation}

\subsubsection{Overall loss}
Finally, the overall unsupervised loss is obtained as
\begin{equation} 
\mathcal{L}=\alpha\mathcal{L}_{po2pl}+\lambda\mathcal{L}_{po2pl},\label{formula3}
\end{equation}
where $\alpha$ and $\lambda$ are balancing factors.

\section{Experiments}
In this section, we first introduce implementation details of our model and benchmark dataset used in our experiments 
and the implementation details of the proposed model. Then, comparing 
to the existing lidar odometry methods, our model can 
obtain competitive results. Finally, we conduct ablation studies to verify the 
effectiveness of the innovative part of our model. 
\subsection{Implementation Details}
The proposed network is implemented in PyTorch \cite{Pytorch} and trained 
with a single NVIDIA Titan RTX GPU. We optimize the parameters with the Adam optimizer
 \cite{Adam} whose hyperparameter values are $\beta_1 = 0.9$, $\beta_2 = 0.99$ and 
$w_{decay} = 10^{-5}$. We adopt step scheduler with a step size of 20 and 
$\gamma = 0.5$ to control the training procedure, the initial learning rate is 
$10^{-4}$ and the 
batch size is 20. The length $S$ of IMU sequence is 15. The maximum horizontal and vertical angle of vertex map 
are $f_w=180^\circ$ and $f_h=23^\circ$, and density of them are $\eta_w = \eta_h = 0.5$. 
The shapes of input maps are $H = 52$ and $W = 720$. The loss weight of formula 
(\ref{formula3}) is set to be $\alpha = 1.0$ and $\lambda = 0.1$. 
The initial side length of voxel downsample is set to 0.3m, it is adjusted 
according to the number of points after downsample, if points are too many, 
we increase the side length, otherwise reduce. The adjustment size is 
0.01m per time. The number of points after downsample is controlled 
within $K\pm 100$ and $K=10240$.
\subsection{Datasets}
The KITTI odometry dataset \cite{KITTI} has 22 different sequences 
with images, 3D lidar point clouds, IMU and other data. Only sequences 00-10 
have an official public ground truth. Among them, only sequence 03 does not 
provide IMU. Therefore, we do not use sequence 03 when there exists the IMU 
assist in our method.

\subsection{Evaluation on the KITTI Dataset}
\begin{figure}[!t]
    \includegraphics[width=\textwidth]{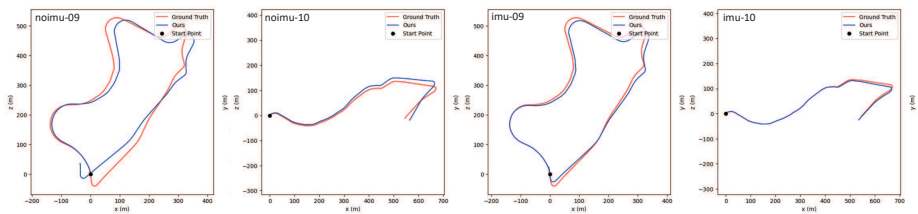}
    \caption{2D estimated trajectories of our method on sequence 09 and 10.} 
    \label{fig3}
\end{figure}

We compare our method with the following methods which can be divided into 
two types. Model-based methods are: 
LOAM \cite{Loam} and LeGO-LOAM \cite{Lego-loam}. 
Learning-based methods are: 
Nubert \emph{et al.} \cite{Nubert}, Cho \emph{et al.} \cite{Cho} and SelfVoxeLO \cite{SelfVoxeLO}. \\
\indent In model-based methods, we show the lidar odometry results of them with mapping and without mapping. \\
\indent In learning-based methods, we use two ways to divide the train and test set. 
First, we use sequences 00-08 for training and 09-10 for testing, 
as Cho \emph{et al.} \cite{Cho} and Nubert \emph{et al.} \cite{Nubert} 
use Sequences 00-08 as their training set. 
We name it as "Ours-easy". Then, we use sequences 00-06 for training and 
07-10 for testing, to compare with SelfVoxeLO which uses Sequences 00-06 as 
training set. We name it as "Ours-hard". \\
\indent Table.~\ref{tab1} contains the details of the results: $t_{rel}$ means 
average translational RMSE (\%) on length of 100m-800m and $r_{rel}$ means 
average rotational RMSE ($^\circ$/100m) on length of 100m-800m. 
LeGO-LOAM is not always more precise by adding imu, 
traditional method is more sensitive to the accuracy of imu (In sequence 00, 
there exists some lack of IMU), 
which is most likely the reason for its accuracy drop. 
Even if the accuracy of the estimation is improved by the IMU, 
the effect is not obvious, especially after the mapping step. 
Our method gains a significant 
improvement by using IMU in test set, and has a certain advantage over traditional method 
without mapping, and is not much lower than with mapping. 
In the easy task (For trajectories results, see Fig.~\ref{fig3}), 
our method without imu assist is also competitive compared to Cho \emph{et al.} \cite{Cho} and Nubert \emph{et al.} \cite{Nubert}
which also project the point clouds into the 2D image coordinate system. 
Our method can acquire a lot of improvements with imu. In the hard task, 
comparing to the most advanced method SelfVoxeLO \cite{SelfVoxeLO} which uses 3D 
convolutions on voxels and consumes much video memory and training time, our method 
also can get comparable results with IMU. Since they 
did not publish the code, we are unable to conduct experiments 
on their method with imu.

\begin{table}[!t] 
  \setlength{\belowcaptionskip}{-0.1cm}
  \caption{KITTI odometry evaluation.}\label{tab1}
  \begin{center}
  \resizebox{1.0\textwidth}{!}{   
    \begin{tabular}{lccccccccccccc}
        \hline
        \multicolumn{1}{l|}{$t_{rel}(\%)$}                        & \multicolumn{1}{c|}{00}                                   & \multicolumn{1}{c|}{01}                                   & \multicolumn{1}{c|}{02}                                   & \multicolumn{1}{c|}{03}                          & \multicolumn{1}{c|}{04}                                   & \multicolumn{1}{c|}{05}                                   & \multicolumn{1}{c|}{06}                                   & \multicolumn{1}{c|}{07}                                   & \multicolumn{1}{c|}{08}                                   & \multicolumn{1}{c|}{09}                                   & \multicolumn{1}{c|}{10}                                   & \multicolumn{1}{c|}{trainavg}                              & testavg                              \\ \hline
        \multicolumn{1}{l|}{LeGO-LOAM(w/ map)\cite{Lego-loam}}    & \multicolumn{1}{c|}{1.44}                                 & \multicolumn{1}{c|}{21.12}                                & \multicolumn{1}{c|}{2.69}                                 & \multicolumn{1}{c|}{1.73}                        & \multicolumn{1}{c|}{1.70}                                 & \multicolumn{1}{c|}{0.98}                                 & \multicolumn{1}{c|}{0.87}                                 & \multicolumn{1}{c|}{0.77}                                 & \multicolumn{1}{c|}{1.35}                                 & \multicolumn{1}{c|}{1.46}                                 & \multicolumn{1}{c|}{1.84}                                 & \multicolumn{2}{c}{3.27}                                                                         \\
        \multicolumn{1}{l|}{LeGO-LOAM(w/ map)+imu}                & \multicolumn{1}{c|}{7.24}                                 & \multicolumn{1}{c|}{20.07}                                & \multicolumn{1}{c|}{2.56}                                 & \multicolumn{1}{c|}{x}                           & \multicolumn{1}{c|}{1.68}                                 & \multicolumn{1}{c|}{0.82}                                 & \multicolumn{1}{c|}{0.86}                                 & \multicolumn{1}{c|}{\textbf{0.67}}                        & \multicolumn{1}{c|}{1.29}                                 & \multicolumn{1}{c|}{1.49}                                 & \multicolumn{1}{c|}{1.75}                                 & \multicolumn{2}{c}{3.84}                                                                         \\
        \multicolumn{1}{l|}{LeGO-LOAM(w/o map)}                   & \multicolumn{1}{c|}{6.98}                                 & \multicolumn{1}{c|}{26.52}                                & \multicolumn{1}{c|}{6.92}                                 & \multicolumn{1}{c|}{6.16}                        & \multicolumn{1}{c|}{3.64}                                 & \multicolumn{1}{c|}{4.57}                                 & \multicolumn{1}{c|}{5.16}                                 & \multicolumn{1}{c|}{4.05}                                 & \multicolumn{1}{c|}{6.01}                                 & \multicolumn{1}{c|}{5.22}                                 & \multicolumn{1}{c|}{7.73}                                 & \multicolumn{2}{c}{7.54}                                                                         \\
        \multicolumn{1}{l|}{LeGO-LOAM(w/o map)+imu}               & \multicolumn{1}{c|}{10.46}                                & \multicolumn{1}{c|}{22.38}                                & \multicolumn{1}{c|}{6.05}                                 & \multicolumn{1}{c|}{x}                           & \multicolumn{1}{c|}{2.04}                                 & \multicolumn{1}{c|}{1.98}                                 & \multicolumn{1}{c|}{2.98}                                 & \multicolumn{1}{c|}{2.99}                                 & \multicolumn{1}{c|}{3.23}                                 & \multicolumn{1}{c|}{3.29}                                 & \multicolumn{1}{c|}{2.74}                                 & \multicolumn{2}{c}{5.81}                                                                         \\
        \multicolumn{1}{l|}{LOAM(w/ map)\cite{Loam}}              & \multicolumn{1}{c|}{\textbf{1.10}}                        & \multicolumn{1}{c|}{\textbf{2.79}}                        & \multicolumn{1}{c|}{\textbf{1.54}}                        & \multicolumn{1}{c|}{\textbf{1.13}}               & \multicolumn{1}{c|}{\textbf{1.45}}                        & \multicolumn{1}{c|}{\textbf{0.75}}                        & \multicolumn{1}{c|}{\textbf{0.72}}                        & \multicolumn{1}{c|}{0.69}                                 & \multicolumn{1}{c|}{\textbf{1.18}}                        & \multicolumn{1}{c|}{\textbf{1.20}}                        & \multicolumn{1}{c|}{\textbf{1.51}}                        & \multicolumn{2}{c}{\textbf{1.28}}                                                                \\
        \multicolumn{1}{l|}{LOAM(w/o map)}                        & \multicolumn{1}{c|}{15.99}                                & \multicolumn{1}{c|}{3.43}                                 & \multicolumn{1}{c|}{9.40}                                 & \multicolumn{1}{c|}{18.18}                       & \multicolumn{1}{c|}{9.59}                                 & \multicolumn{1}{c|}{9.16}                                 & \multicolumn{1}{c|}{8.91}                                 & \multicolumn{1}{c|}{10.87}                                & \multicolumn{1}{c|}{12.72}                                & \multicolumn{1}{c|}{8.10}                                 & \multicolumn{1}{c|}{12.67}                                & \multicolumn{2}{c}{10.82}                                                                        \\ \hline
        \multicolumn{1}{l|}{Nubert et al.\cite{Nubert}}           & \multicolumn{1}{c|}{NA}                                   & \multicolumn{1}{c|}{NA}                                   & \multicolumn{1}{c|}{NA}                                   & \multicolumn{1}{c|}{NA}                          & \multicolumn{1}{c|}{NA}                                   & \multicolumn{1}{c|}{NA}                                   & \multicolumn{1}{c|}{NA}                                   & \multicolumn{1}{c|}{NA}                                   & \multicolumn{1}{c|}{NA}                                   & \multicolumn{1}{c|}{6.05}                                 & \multicolumn{1}{c|}{6.44}                                 & \multicolumn{1}{c|}{3.00}                                 & 6.25                                 \\
        \multicolumn{1}{l|}{Cho et al.\cite{Cho}}                 & \multicolumn{1}{c|}{NA}                                   & \multicolumn{1}{c|}{NA}                                   & \multicolumn{1}{c|}{NA}                                   & \multicolumn{1}{c|}{NA}                          & \multicolumn{1}{c|}{NA}                                   & \multicolumn{1}{c|}{NA}                                   & \multicolumn{1}{c|}{NA}                                   & \multicolumn{1}{c|}{NA}                                   & \multicolumn{1}{c|}{NA}                                   & \multicolumn{1}{c|}{4.87}                                 & \multicolumn{1}{c|}{5.02}                                 & \multicolumn{1}{c|}{3.68}                                 & 4.95                                 \\
        \multicolumn{1}{l|}{{\color[HTML]{2F75B5} Ours-easy}}     & \multicolumn{1}{c|}{{\color[HTML]{2F75B5} \textbf{1.33}}} & \multicolumn{1}{c|}{{\color[HTML]{2F75B5} \textbf{3.40}}} & \multicolumn{1}{c|}{{\color[HTML]{2F75B5} 1.53}}          & \multicolumn{1}{c|}{{\color[HTML]{2F75B5} 1.43}} & \multicolumn{1}{c|}{{\color[HTML]{2F75B5} 1.26}}          & \multicolumn{1}{c|}{{\color[HTML]{2F75B5} 1.22}}          & \multicolumn{1}{c|}{{\color[HTML]{2F75B5} 1.19}}          & \multicolumn{1}{c|}{{\color[HTML]{2F75B5} \textbf{0.97}}} & \multicolumn{1}{c|}{{\color[HTML]{2F75B5} 1.92}}          & \multicolumn{1}{c|}{{\color[HTML]{2F75B5} 3.87}}          & \multicolumn{1}{c|}{{\color[HTML]{2F75B5} 2.69}}          & \multicolumn{1}{c|}{{\color[HTML]{2F75B5} 1.58}}          & {\color[HTML]{2F75B5} 3.28}          \\
        \multicolumn{1}{l|}{{\color[HTML]{2F75B5} Ours-easy+imu}} & \multicolumn{1}{c|}{{\color[HTML]{2F75B5} 1.50}}          & \multicolumn{1}{c|}{{\color[HTML]{2F75B5} 3.44}}          & \multicolumn{1}{c|}{{\color[HTML]{2F75B5} \textbf{1.33}}} & \multicolumn{1}{c|}{{\color[HTML]{2F75B5} x}}    & \multicolumn{1}{c|}{{\color[HTML]{2F75B5} \textbf{0.94}}} & \multicolumn{1}{c|}{{\color[HTML]{2F75B5} \textbf{0.98}}} & \multicolumn{1}{c|}{{\color[HTML]{2F75B5} \textbf{0.90}}} & \multicolumn{1}{c|}{{\color[HTML]{2F75B5} 1.00}}          & \multicolumn{1}{c|}{{\color[HTML]{2F75B5} \textbf{1.63}}} & \multicolumn{1}{c|}{{\color[HTML]{2F75B5} \textbf{2.24}}} & \multicolumn{1}{c|}{{\color[HTML]{2F75B5} \textbf{1.83}}} & \multicolumn{1}{c|}{{\color[HTML]{2F75B5} \textbf{1.46}}} & {\color[HTML]{2F75B5} \textbf{2.03}} \\ \hline
        \multicolumn{1}{l|}{SelfVoxelLO\cite{SelfVoxeLO}}         & \multicolumn{1}{c|}{NA}                                   & \multicolumn{1}{c|}{NA}                                   & \multicolumn{1}{c|}{NA}                                   & \multicolumn{1}{c|}{NA}                          & \multicolumn{1}{c|}{NA}                                   & \multicolumn{1}{c|}{NA}                                   & \multicolumn{1}{c|}{NA}                                   & \multicolumn{1}{c|}{\textbf{3.09}}                        & \multicolumn{1}{c|}{\textbf{3.16}}                        & \multicolumn{1}{c|}{3.01}                                 & \multicolumn{1}{c|}{3.48}                                 & \multicolumn{1}{c|}{2.50}                                 & \textbf{3.19}                        \\
        \multicolumn{1}{l|}{{\color[HTML]{0070C0} Ours-hard}}     & \multicolumn{1}{c|}{{\color[HTML]{0070C0} 1.58}}          & \multicolumn{1}{c|}{{\color[HTML]{0070C0} \textbf{3.42}}} & \multicolumn{1}{c|}{{\color[HTML]{0070C0} 2.27}}          & \multicolumn{1}{c|}{{\color[HTML]{0070C0} 2.53}} & \multicolumn{1}{c|}{{\color[HTML]{0070C0} 0.96}}          & \multicolumn{1}{c|}{{\color[HTML]{0070C0} 1.36}}          & \multicolumn{1}{c|}{{\color[HTML]{0070C0} \textbf{0.99}}} & \multicolumn{1}{c|}{{\color[HTML]{0070C0} 6.58}}          & \multicolumn{1}{c|}{{\color[HTML]{0070C0} 6.89}}          & \multicolumn{1}{c|}{{\color[HTML]{0070C0} 5.77}}          & \multicolumn{1}{c|}{{\color[HTML]{0070C0} 4.04}}          & \multicolumn{1}{c|}{{\color[HTML]{0070C0} 1.87}}          & {\color[HTML]{0070C0} 5.82}          \\
        \multicolumn{1}{l|}{{\color[HTML]{0070C0} Ours-hard+imu}} & \multicolumn{1}{c|}{{\color[HTML]{0070C0} \textbf{1.38}}} & \multicolumn{1}{c|}{{\color[HTML]{0070C0} 3.46}}          & \multicolumn{1}{c|}{{\color[HTML]{0070C0} \textbf{1.42}}} & \multicolumn{1}{c|}{{\color[HTML]{0070C0} x}}    & \multicolumn{1}{c|}{{\color[HTML]{0070C0} \textbf{0.98}}} & \multicolumn{1}{c|}{{\color[HTML]{0070C0} \textbf{1.26}}} & \multicolumn{1}{c|}{{\color[HTML]{0070C0} 0.94}}          & \multicolumn{1}{c|}{{\color[HTML]{0070C0} 3.45}}          & \multicolumn{1}{c|}{{\color[HTML]{0070C0} 4.05}}          & \multicolumn{1}{c|}{{\color[HTML]{0070C0} \textbf{2.77}}} & \multicolumn{1}{c|}{{\color[HTML]{0070C0} \textbf{2.16}}} & \multicolumn{1}{c|}{{\color[HTML]{0070C0} \textbf{1.57}}} & {\color[HTML]{0070C0} \textbf{3.11}} \\ \hline \hline
        \multicolumn{1}{l|}{$r_{rel}(^\circ/100m)$}               & \multicolumn{1}{c|}{00}                                   & \multicolumn{1}{c|}{01}                                   & \multicolumn{1}{c|}{02}                                   & \multicolumn{1}{c|}{03}                          & \multicolumn{1}{c|}{04}                                   & \multicolumn{1}{c|}{05}                                   & \multicolumn{1}{c|}{06}                                   & \multicolumn{1}{c|}{07}                                   & \multicolumn{1}{c|}{08}                                   & \multicolumn{1}{c|}{09}                                   & \multicolumn{1}{c|}{10}                                   & \multicolumn{1}{c|}{trainavg}                              & testavg                              \\ \hline
        \multicolumn{1}{l|}{LeGO-LOAM(w/ map)}                    & \multicolumn{1}{c|}{0.65}                                 & \multicolumn{1}{c|}{2.17}                                 & \multicolumn{1}{c|}{0.99}                                 & \multicolumn{1}{c|}{0.99}                        & \multicolumn{1}{c|}{0.69}                                 & \multicolumn{1}{c|}{0.47}                                 & \multicolumn{1}{c|}{0.45}                                 & \multicolumn{1}{c|}{0.51}                                 & \multicolumn{1}{c|}{0.58}                                 & \multicolumn{1}{c|}{0.64}                                 & \multicolumn{1}{c|}{0.74}                                 & \multicolumn{2}{c}{0.81}                                                                         \\
        \multicolumn{1}{l|}{LeGO-LOAM(w/ map)+imu}                & \multicolumn{1}{c|}{2.44}                                 & \multicolumn{1}{c|}{0.61}                                 & \multicolumn{1}{c|}{0.91}                                 & \multicolumn{1}{c|}{x}                           & \multicolumn{1}{c|}{0.59}                                 & \multicolumn{1}{c|}{\textbf{0.38}}                        & \multicolumn{1}{c|}{0.43}                                 & \multicolumn{1}{c|}{0.38}                                 & \multicolumn{1}{c|}{0.53}                                 & \multicolumn{1}{c|}{0.58}                                 & \multicolumn{1}{c|}{0.63}                                 & \multicolumn{2}{c}{0.75}                                                                         \\
        \multicolumn{1}{l|}{LeGO-LOAM(w/o map)}                   & \multicolumn{1}{c|}{3.27}                                 & \multicolumn{1}{c|}{4.61}                                 & \multicolumn{1}{c|}{3.10}                                 & \multicolumn{1}{c|}{3.42}                        & \multicolumn{1}{c|}{2.98}                                 & \multicolumn{1}{c|}{2.38}                                 & \multicolumn{1}{c|}{2.24}                                 & \multicolumn{1}{c|}{2.41}                                 & \multicolumn{1}{c|}{2.85}                                 & \multicolumn{1}{c|}{2.61}                                 & \multicolumn{1}{c|}{4.03}                                 & \multicolumn{2}{c}{3.08}                                                                         \\
        \multicolumn{1}{l|}{LeGO-LOAM(w/o map)+imu}               & \multicolumn{1}{c|}{3.72}                                 & \multicolumn{1}{c|}{1.79}                                 & \multicolumn{1}{c|}{2.12}                                 & \multicolumn{1}{c|}{x}                           & \multicolumn{1}{c|}{0.88}                                 & \multicolumn{1}{c|}{0.88}                                 & \multicolumn{1}{c|}{1.24}                                 & \multicolumn{1}{c|}{1.64}                                 & \multicolumn{1}{c|}{1.23}                                 & \multicolumn{1}{c|}{1.75}                                 & \multicolumn{1}{c|}{1.57}                                 & \multicolumn{2}{c}{1.68}                                                                         \\
        \multicolumn{1}{l|}{LOAM(w/ map)}                         & \multicolumn{1}{c|}{\textbf{0.53}}                        & \multicolumn{1}{c|}{\textbf{0.55}}                        & \multicolumn{1}{c|}{\textbf{0.55}}                        & \multicolumn{1}{c|}{\textbf{0.65}}               & \multicolumn{1}{c|}{\textbf{0.50}}                        & \multicolumn{1}{c|}{\textbf{0.38}}                        & \multicolumn{1}{c|}{\textbf{0.39}}                        & \multicolumn{1}{c|}{\textbf{0.50}}                        & \multicolumn{1}{c|}{\textbf{0.44}}                        & \multicolumn{1}{c|}{\textbf{0.48}}                        & \multicolumn{1}{c|}{\textbf{0.57}}                        & \multicolumn{2}{c}{\textbf{0.50}}                                                                \\
        \multicolumn{1}{l|}{LOAM(w/o map)}                        & \multicolumn{1}{c|}{6.25}                                 & \multicolumn{1}{c|}{0.93}                                 & \multicolumn{1}{c|}{3.68}                                 & \multicolumn{1}{c|}{9.91}                        & \multicolumn{1}{c|}{4.57}                                 & \multicolumn{1}{c|}{4.10}                                 & \multicolumn{1}{c|}{4.63}                                 & \multicolumn{1}{c|}{6.76}                                 & \multicolumn{1}{c|}{5.77}                                 & \multicolumn{1}{c|}{4.30}                                 & \multicolumn{1}{c|}{8.79}                                 & \multicolumn{2}{c}{5.43}                                                                         \\ \hline
        \multicolumn{1}{l|}{Nubert  et al.}                       & \multicolumn{1}{c|}{NA}                                   & \multicolumn{1}{c|}{NA}                                   & \multicolumn{1}{c|}{NA}                                   & \multicolumn{1}{c|}{NA}                          & \multicolumn{1}{c|}{NA}                                   & \multicolumn{1}{c|}{NA}                                   & \multicolumn{1}{c|}{NA}                                   & \multicolumn{1}{c|}{NA}                                   & \multicolumn{1}{c|}{NA}                                   & \multicolumn{1}{c|}{2.15}                                 & \multicolumn{1}{c|}{3.00}                                 & \multicolumn{1}{c|}{1.38}                                 & 2.58                                 \\
        \multicolumn{1}{l|}{Cho et al.}                           & \multicolumn{1}{c|}{NA}                                   & \multicolumn{1}{c|}{NA}                                   & \multicolumn{1}{c|}{NA}                                   & \multicolumn{1}{c|}{NA}                          & \multicolumn{1}{c|}{NA}                                   & \multicolumn{1}{c|}{NA}                                   & \multicolumn{1}{c|}{NA}                                   & \multicolumn{1}{c|}{NA}                                   & \multicolumn{1}{c|}{NA}                                   & \multicolumn{1}{c|}{1.95}                                 & \multicolumn{1}{c|}{1.83}                                 & \multicolumn{1}{c|}{0.87}                                 & 1.89                                 \\
        \multicolumn{1}{l|}{{\color[HTML]{2F75B5} Ours-easy}}     & \multicolumn{1}{c|}{{\color[HTML]{2F75B5} \textbf{0.69}}} & \multicolumn{1}{c|}{{\color[HTML]{2F75B5} \textbf{0.97}}} & \multicolumn{1}{c|}{{\color[HTML]{2F75B5} 0.68}}          & \multicolumn{1}{c|}{{\color[HTML]{2F75B5} 1.04}} & \multicolumn{1}{c|}{{\color[HTML]{2F75B5} \textbf{0.73}}} & \multicolumn{1}{c|}{{\color[HTML]{2F75B5} 0.66}}          & \multicolumn{1}{c|}{{\color[HTML]{2F75B5} 0.64}}          & \multicolumn{1}{c|}{{\color[HTML]{2F75B5} 0.58}}          & \multicolumn{1}{c|}{{\color[HTML]{2F75B5} \textbf{0.78}}} & \multicolumn{1}{c|}{{\color[HTML]{2F75B5} 1.67}}          & \multicolumn{1}{c|}{{\color[HTML]{2F75B5} 1.97}}          & \multicolumn{1}{c|}{{\color[HTML]{2F75B5} 0.75}}          & {\color[HTML]{2F75B5} 1.82}          \\
        \multicolumn{1}{l|}{{\color[HTML]{2F75B5} Ours-easy+imu}} & \multicolumn{1}{c|}{{\color[HTML]{2F75B5} 0.70}}          & \multicolumn{1}{c|}{{\color[HTML]{2F75B5} 0.99}}          & \multicolumn{1}{c|}{{\color[HTML]{2F75B5} \textbf{0.59}}} & \multicolumn{1}{c|}{{\color[HTML]{2F75B5} x}}    & \multicolumn{1}{c|}{{\color[HTML]{2F75B5} 0.78}}          & \multicolumn{1}{c|}{{\color[HTML]{2F75B5} \textbf{0.56}}} & \multicolumn{1}{c|}{{\color[HTML]{2F75B5} \textbf{0.45}}} & \multicolumn{1}{c|}{{\color[HTML]{2F75B5} \textbf{0.54}}} & \multicolumn{1}{c|}{{\color[HTML]{2F75B5} \textbf{0.78}}} & \multicolumn{1}{c|}{{\color[HTML]{2F75B5} \textbf{1.13}}} & \multicolumn{1}{c|}{{\color[HTML]{2F75B5} \textbf{1.14}}} & \multicolumn{1}{c|}{{\color[HTML]{2F75B5} \textbf{0.67}}} & {\color[HTML]{2F75B5} \textbf{1.14}} \\ \hline
        \multicolumn{1}{l|}{SelfVoxelLO}                          & \multicolumn{1}{c|}{NA}                                   & \multicolumn{1}{c|}{NA}                                   & \multicolumn{1}{c|}{NA}                                   & \multicolumn{1}{c|}{NA}                          & \multicolumn{1}{c|}{NA}                                   & \multicolumn{1}{c|}{NA}                                   & \multicolumn{1}{c|}{NA}                                   & \multicolumn{1}{c|}{\textbf{1.81}}                        & \multicolumn{1}{c|}{\textbf{1.14}}                        & \multicolumn{1}{c|}{1.14}                                 & \multicolumn{1}{c|}{\textbf{1.11}}                        & \multicolumn{1}{c|}{1.11}                                 & \textbf{1.30}                        \\
        \multicolumn{1}{l|}{{\color[HTML]{0070C0} Ours-hard}}     & \multicolumn{1}{c|}{{\color[HTML]{0070C0} 0.91}}          & \multicolumn{1}{c|}{{\color[HTML]{0070C0} 1.09}}          & \multicolumn{1}{c|}{{\color[HTML]{0070C0} 1.19}}          & \multicolumn{1}{c|}{{\color[HTML]{0070C0} 1.42}} & \multicolumn{1}{c|}{{\color[HTML]{0070C0} \textbf{0.61}}} & \multicolumn{1}{c|}{{\color[HTML]{0070C0} 0.78}}          & \multicolumn{1}{c|}{{\color[HTML]{0070C0} 0.64}}          & \multicolumn{1}{c|}{{\color[HTML]{0070C0} 4.56}}          & \multicolumn{1}{c|}{{\color[HTML]{0070C0} 2.86}}          & \multicolumn{1}{c|}{{\color[HTML]{0070C0} 2.34}}          & \multicolumn{1}{c|}{{\color[HTML]{0070C0} 2.89}}          & \multicolumn{1}{c|}{{\color[HTML]{0070C0} 0.95}}          & {\color[HTML]{0070C0} 3.16}          \\
        \multicolumn{1}{l|}{{\color[HTML]{0070C0} Ours-hard+imu}} & \multicolumn{1}{c|}{{\color[HTML]{0070C0} \textbf{0.62}}} & \multicolumn{1}{c|}{{\color[HTML]{0070C0} \textbf{0.98}}} & \multicolumn{1}{c|}{{\color[HTML]{0070C0} \textbf{0.67}}} & \multicolumn{1}{c|}{{\color[HTML]{0070C0} x}}    & \multicolumn{1}{c|}{{\color[HTML]{0070C0} 0.67}}          & \multicolumn{1}{c|}{{\color[HTML]{0070C0} \textbf{0.64}}} & \multicolumn{1}{c|}{{\color[HTML]{0070C0} \textbf{0.56}}} & \multicolumn{1}{c|}{{\color[HTML]{0070C0} 2.17}}          & \multicolumn{1}{c|}{{\color[HTML]{0070C0} 1.63}}          & \multicolumn{1}{c|}{{\color[HTML]{0070C0} \textbf{1.25}}} & \multicolumn{1}{c|}{{\color[HTML]{0070C0} \textbf{1.11}}} & \multicolumn{1}{c|}{{\color[HTML]{0070C0} \textbf{0.69}}} & {\color[HTML]{0070C0} 1.54}          \\ \hline
        \multicolumn{14}{l}{NA: The result of other papers do not provide.}                                                                                                                                                                                                                                                                                                                                                                                                                                                                                                                                                                                                                                                                                                                                                                     \\
        \multicolumn{14}{l}{x: Do not use this sequence in method.}                                                                                                                                                                                                                                                                                                                                                                                                                                                                                                                                                                                                                                                                                                                                                                             \\
        \multicolumn{14}{l}{Trainavg and testavg of traditional methods are the average results of all 00-10 sequences.}                                                                                                                                                                                                                                                                                                                                                                                                                                                                                                                                                                                                                                                                                                                           
        \end{tabular}} 
  \end{center}
\end{table}

\subsection{Ablation Study}
\subsubsection{IMU}
\begin{figure}[!t]
    \centering
    \includegraphics[width=9cm]{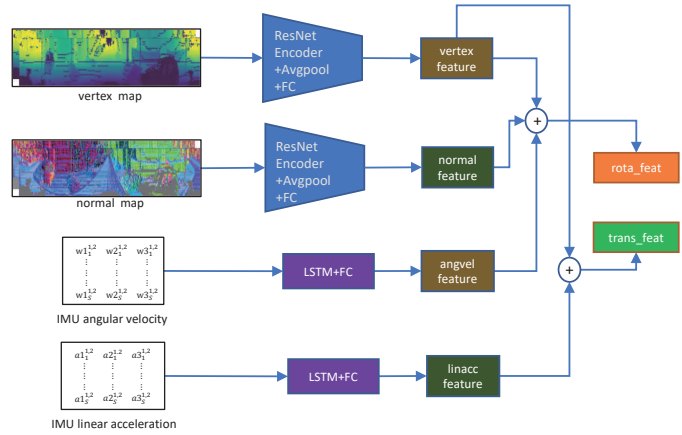}
    \caption{Use IMU only as feature.} 
    \label{fig4}
\end{figure}
\begin{table}[!t]
    \setlength{\belowcaptionskip}{-0.1cm}
    \caption{Comparison among different ways to preprocess imu and whether using imu.}\label{tab2}
     \begin{center}
      \resizebox{1.0\textwidth}{!}{   
        \begin{tabular}{l|c|c|c|c|c|c|c|c|c|c|c|c|c}
            \hline
            $t_{rel}(\%)$                            & 00                          & 01                                   & 02                                   & 03                       & 04                                   & 05                                   & 06                                   & 07                                   & 08                          & 09                                   & 10                                   & trainavg                              & testavg                              \\ \hline
            {\color[HTML]{2F75B5} imu(w preprocess)} & {\color[HTML]{2F75B5} 1.50} & {\color[HTML]{2F75B5} 3.44}          & {\color[HTML]{2F75B5} \textbf{1.33}} & {\color[HTML]{2F75B5} x} & {\color[HTML]{2F75B5} \textbf{0.94}} & {\color[HTML]{2F75B5} \textbf{0.98}} & {\color[HTML]{2F75B5} \textbf{0.90}} & {\color[HTML]{2F75B5} 1.00}          & {\color[HTML]{2F75B5} 1.63} & {\color[HTML]{2F75B5} \textbf{2.24}} & {\color[HTML]{2F75B5} \textbf{1.83}} & {\color[HTML]{2F75B5} \textbf{1.58}} & {\color[HTML]{2F75B5} \textbf{2.03}} \\
            imu(w/o preprocess)                    & 1.35                        & 3.56                                 & 1.57                                 & x                        & 1.07                                 & 1.21                                 & 1.03                                 & \textbf{0.90}                        & \textbf{1.59}               & 2.46                                 & 1.87                                 & 1.66                                 & 2.17                                 \\
            noimu                                  & \textbf{1.33}               & \textbf{3.40}                        & 1.53                                 & 1.43                     & 1.26                                 & 1.22                                 & 1.19                                 & 0.97                                 & 1.92                        & 3.87                                 & 2.69                                 & 1.89                                 & 3.28                                 \\ \hline \hline
            $r_{rel}(^\circ/100m)$                   & 00                          & 01                                   & 02                                   & 03                       & 04                                   & 05                                   & 06                                   & 07                                   & 08                          & 09                                   & 10                                   & trainavg                              & testavg                              \\ \hline
            {\color[HTML]{2F75B5} imu(w preprocess)}& {\color[HTML]{2F75B5} 0.70} & {\color[HTML]{2F75B5} \textbf{0.99}} & {\color[HTML]{2F75B5} \textbf{0.59}} & {\color[HTML]{2F75B5} x} & {\color[HTML]{2F75B5} 0.78}          & {\color[HTML]{2F75B5} \textbf{0.56}} & {\color[HTML]{2F75B5} \textbf{0.45}} & {\color[HTML]{2F75B5} \textbf{0.54}} & {\color[HTML]{2F75B5} 0.78} & {\color[HTML]{2F75B5} \textbf{1.13}} & {\color[HTML]{2F75B5} \textbf{1.14}} & {\color[HTML]{2F75B5} \textbf{0.76}} & {\color[HTML]{2F75B5} \textbf{1.14}} \\
            imu(w/o preprocess)                    & \textbf{0.67}               & 1.01                                 & 0.69                                 & x                        & \textbf{0.63}                        & 0.68                                 & 0.56                                 & 0.60                                 & \textbf{0.71}               & \textbf{1.13}                        & 1.28                                 & 0.80                                 & 1.20                                 \\
            noimu                                  & 0.69                        & 0.97                                 & 0.68                                 & 1.04                     & 0.73                                 & 0.66                                 & 0.64                                 & 0.58                                 & 0.78                        & 1.67                                 & 1.97                                 & 0.95                                 & 1.82                                 \\ \hline
        \end{tabular}} 
    \end{center}
\end{table}
As mentioned earlier, IMU can greatly improve the accuracy of odometry, 
but the role played by different IMU utilization methods is also different. 
If only use IMU to extract 
features through the network, and directly merge with the feature of the point 
clouds, the effect is limited (see Fig.~\ref{fig4}).
Our method uses IMU and LSTM network 
to estimate a relative initial pose, project vertex 
image and normal vector image of the original current frame, and then send the projection images into the point 
clouds feature extraction network, so that the IMU can not only have a direct 
connection with the final odometry estimate network, but also make the 
coordinate of two consecutive frames closer. 
The comparison is shown in Table.~\ref{tab2}.

\subsubsection{Different operations to obtain the rotation and translation features}
\begin{figure}[!t]
    \centering
    \includegraphics[width=6cm]{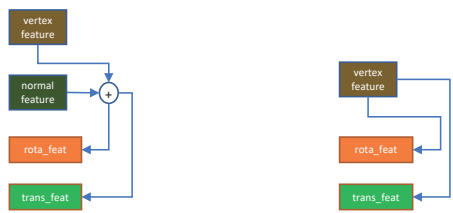}
    \caption{The network structure of learning translation and rotation features from concatenated vetex and normal features simultaneously (left) and the network structure
    without the normal feature (right).} 
    \label{fig5}
\end{figure}
\begin{table}
    \setlength{\belowcaptionskip}{-0.1cm}
    \caption{Comparison among whether distinguishing features (dist) and whether using normal.}\label{tab3}
    \begin{center}
    \resizebox{1.0\textwidth}{!}{   
        \begin{tabular}{l|c|c|c|c|c|c|c|c|c|c|c|c|c}
            \hline
            $t_{rel}(\%)$                                         & 00                                   & 01                                   & 02                                   & 03                                   & 04                          & 05                                   & 06                                   & 07                                   & 08                                   & 09                                   & 10                                   & trainavg                              & testavg                              \\ \hline
            {\color[HTML]{2F75B5} imu(w normal, w dist)}   & {\color[HTML]{2F75B5} 1.50}          & {\color[HTML]{2F75B5} \textbf{3.44}} & {\color[HTML]{2F75B5} \textbf{1.33}} & {\color[HTML]{2F75B5} x}             & {\color[HTML]{2F75B5} 0.94} & {\color[HTML]{2F75B5} \textbf{0.98}} & {\color[HTML]{2F75B5} \textbf{0.90}} & {\color[HTML]{2F75B5} 1.00}          & {\color[HTML]{2F75B5} \textbf{1.63}} & {\color[HTML]{2F75B5} \textbf{2.24}} & {\color[HTML]{2F75B5} \textbf{1.83}} & {\color[HTML]{2F75B5} \textbf{1.58}} & {\color[HTML]{2F75B5} \textbf{2.03}} \\
            imu(w normal, w/o dist)                        & \textbf{1.45}                        & 3.68                                 & 2.03                                 & x                                    & \textbf{0.72}               & 1.11                                 & 1.15                                 & \textbf{0.68}                        & 1.67                                 & 3.44                                 & 1.86                                 & 1.78                                 & 2.65                                 \\
            imu(w/o normal, w/o dist)                      & 2.54                                 & 3.81                                 & 4.13                                 & x                                    & 0.95                        & 1.77                                 & 0.99                                 & 1.25                                 & 1.93                                 & 2.72                                 & 2.21                                 & 2.23                                 & 2.47                                 \\ \hline
            {\color[HTML]{2F75B5} noimu(w normal, w dist)} & {\color[HTML]{2F75B5} \textbf{1.33}} & {\color[HTML]{2F75B5} \textbf{3.40}} & {\color[HTML]{2F75B5} \textbf{1.53}} & {\color[HTML]{2F75B5} \textbf{1.43}} & {\color[HTML]{2F75B5} 1.26} & {\color[HTML]{2F75B5} 1.22}          & {\color[HTML]{2F75B5} 1.19}          & {\color[HTML]{2F75B5} \textbf{0.97}} & {\color[HTML]{2F75B5} 1.92}          & {\color[HTML]{2F75B5} \textbf{3.87}} & {\color[HTML]{2F75B5} \textbf{2.69}} & {\color[HTML]{2F75B5} \textbf{1.89}} & {\color[HTML]{2F75B5} \textbf{3.28}} \\
            noimu(w normal, w/o dist)                      & 1.49                                 & 3.95                                 & 2.49                                 & 2.27                                 & \textbf{0.88}               & \textbf{1.19}                        & \textbf{0.90}                        & 1.47                                 & 2.02                                 & 4.93                                 & 4.34                                 & 2.36                                 & 4.64                                 \\
            noimu(w/o normal, w/o dist)                    & 1.63                                 & 4.96                                 & 2.99                                 & 2.36                                 & 2.15                        & 1.31                                 & 1.31                                 & 1.51                                 & \textbf{1.89}                        & 5.75                                 & 6.11                                 & 2.91                                 & 5.93                                 \\ \hline \hline
            $r_{rel}(^\circ/100m)$                                & 00                                   & 01                                   & 02                                   & 03                                   & 04                          & 05                                   & 06                                   & 07                                   & 08                                   & 09                                   & 10                                   & trainavg                              & testavg                              \\ \hline
            {\color[HTML]{2F75B5} imu(w normal, w dist)}   & {\color[HTML]{2F75B5} \textbf{0.70}} & {\color[HTML]{2F75B5} \textbf{0.99}} & {\color[HTML]{2F75B5} \textbf{0.59}} & {\color[HTML]{2F75B5} x}             & {\color[HTML]{2F75B5} 0.78} & {\color[HTML]{2F75B5} \textbf{0.56}} & {\color[HTML]{2F75B5} \textbf{0.45}} & {\color[HTML]{2F75B5} 0.54}          & {\color[HTML]{2F75B5} 0.78}          & {\color[HTML]{2F75B5} \textbf{1.13}} & {\color[HTML]{2F75B5} \textbf{1.14}} & {\color[HTML]{2F75B5} \textbf{0.76}} & {\color[HTML]{2F75B5} \textbf{1.14}} \\
            imu(w normal, w/o dist)                        & 0.65                                 & 1.04                                 & 0.96                                 & x                                    & \textbf{0.53}               & \textbf{0.56}                        & 0.58                                 & \textbf{0.46}                        & \textbf{0.64}                        & 1.45                                 & 1.15                                 & 0.80                                 & 1.30                                 \\
            imu(w/o normal, w/o dist)                      & 1.31                                 & 1.05                                 & 1.60                                 & x                                    & 0.52                        & 0.88                                 & 0.48                                 & 0.87                                 & 0.93                                 & 1.15                                 & 1.21                                 & 1.00                                 & 1.18                                 \\ \hline
            {\color[HTML]{2F75B5} noimu(w normal, w dist)} & {\color[HTML]{2F75B5} \textbf{0.69}} & {\color[HTML]{2F75B5} \textbf{0.97}} & {\color[HTML]{2F75B5} \textbf{0.68}} & {\color[HTML]{2F75B5} \textbf{1.04}} & {\color[HTML]{2F75B5} 0.73} & {\color[HTML]{2F75B5} \textbf{0.66}} & {\color[HTML]{2F75B5} \textbf{0.64}} & {\color[HTML]{2F75B5} \textbf{0.58}} & {\color[HTML]{2F75B5} \textbf{0.78}} & {\color[HTML]{2F75B5} \textbf{1.67}} & {\color[HTML]{2F75B5} \textbf{1.97}} & {\color[HTML]{2F75B5} \textbf{0.95}} & {\color[HTML]{2F75B5} \textbf{1.82}} \\
            noimu(w normal, w/o dist)                      & 0.88                                 & 1.24                                 & 1.20                                 & 1.38                                 & \textbf{0.66}               & 0.70                                 & 0.58                                 & 1.03                                 & 0.95                                 & 1.92                                 & 2.06                                 & 1.15                                 & 1.99                                 \\
            noimu(w/o normal, w/o dist)                    & 0.90                                 & 1.48                                 & 1.37                                 & 1.49                                 & 1.38                        & 0.79                                 & 0.73                                 & 1.08                                 & 0.93                                 & 2.31                                 & 2.73                                 & 1.38                                 & 2.52                                 \\ \hline
        \end{tabular}} 
    \end{center}
\end{table}
The normal vector contains the relationship between a point and its 
surrounding points, and can be used as feature of pose estimation 
just like the point itself. Through the calculation formula 
of the normal vector, we can know that the change of the normal vector 
is only related to the orientation, and the 
translation will not bring about the change of the normal vector. 
Therefore, we only use the feature of the point to estimate the 
translation. We compare the original method with the 
two strategies of not using normal vectors as the network 
input and not distinguishing feature of the normals and points (see Fig.~\ref{fig5}). The comparison is shown in Table.~\ref{tab3}.

\subsubsection{Attention}
\begin{table}[!t]
    \setlength{\belowcaptionskip}{-0.1cm}
    \caption{Comparison among whether using attention module.}\label{tab4}
    \centering
    \begin{center}
    \resizebox{1.0\textwidth}{!}{   
        \begin{tabular}{l|c|c|c|c|c|c|c|c|c|c|c|c|c}
            \hline
            $t_{rel}(\%)$                             & 00                                   & 01                                   & 02                                   & 03                                   & 04                                   & 05                                   & 06                          & 07                                   & 08                                   & 09                                   & 10                                   & trainavg                              & testavg                              \\ \hline
            {\color[HTML]{2F75B5} imu(w attention)}   & {\color[HTML]{2F75B5} 1.50}          & {\color[HTML]{2F75B5} \textbf{3.44}} & {\color[HTML]{2F75B5} \textbf{1.33}} & {\color[HTML]{2F75B5} x}             & {\color[HTML]{2F75B5} 0.94}          & {\color[HTML]{2F75B5} 0.98}          & {\color[HTML]{2F75B5} 0.90} & {\color[HTML]{2F75B5} 1.00}          & {\color[HTML]{2F75B5} 1.63}          & {\color[HTML]{2F75B5} \textbf{2.24}} & {\color[HTML]{2F75B5} 1.83}          & {\color[HTML]{2F75B5} 1.58}          & {\color[HTML]{2F75B5} \textbf{2.03}} \\
            imu(w fc+activation)                      & \textbf{1.19}                        & 3.49                                 & 1.48                                 & x                                    & \textbf{0.83}                        & \textbf{0.95}                        & \textbf{0.64}               & \textbf{0.91}                        & \textbf{1.49}                        & 3.21                                 & \textbf{1.54}                        & \textbf{1.57}                        & 2.38                                 \\ \hline
            {\color[HTML]{2F75B5} noimu(w attention)} & {\color[HTML]{2F75B5} \textbf{1.33}} & {\color[HTML]{2F75B5} \textbf{3.40}} & {\color[HTML]{2F75B5} \textbf{1.53}} & {\color[HTML]{2F75B5} \textbf{1.43}} & {\color[HTML]{2F75B5} 1.26}          & {\color[HTML]{2F75B5} \textbf{1.22}} & {\color[HTML]{2F75B5} 1.19} & {\color[HTML]{2F75B5} \textbf{0.97}} & {\color[HTML]{2F75B5} 1.92}          & {\color[HTML]{2F75B5} \textbf{3.87}} & {\color[HTML]{2F75B5} \textbf{2.69}} & {\color[HTML]{2F75B5} \textbf{1.89}} & {\color[HTML]{2F75B5} \textbf{3.28}} \\
            noimu(w fc+activation)                    & 1.65                                 & 3.59                                 & 1.67                                 & 1.88                                 & \textbf{0.87}                        & 1.34                                 & \textbf{1.10}               & 1.23                                 & \textbf{1.76}                        & 6.64                                 & 3.25                                 & 2.27                                 & 4.95                                 \\ \hline \hline
            $r_{rel}(^\circ/100m)$                    & 00                                   & 01                                   & 02                                   & 03                                   & 04                                   & 05                                   & 06                          & 07                                   & 08                                   & 09                                   & 10                                   & trainavg                              & testavg                              \\ \hline
            {\color[HTML]{2F75B5} imu(w attention)}   & {\color[HTML]{2F75B5} 0.70}          & {\color[HTML]{2F75B5} 0.99}          & {\color[HTML]{2F75B5} \textbf{0.59}} & {\color[HTML]{2F75B5} x}             & {\color[HTML]{2F75B5} \textbf{0.78}} & {\color[HTML]{2F75B5} 0.56}          & {\color[HTML]{2F75B5} 0.45} & {\color[HTML]{2F75B5} \textbf{0.54}} & {\color[HTML]{2F75B5} 0.78}          & {\color[HTML]{2F75B5} \textbf{1.13}} & {\color[HTML]{2F75B5} 1.14}          & {\color[HTML]{2F75B5} \textbf{0.76}} & {\color[HTML]{2F75B5} 1.14}          \\
            imu(w fc+activation)                      & \textbf{0.62}                        & \textbf{0.97}                        & 0.64                                 & x                                    & 1.02                                 & \textbf{0.54}                        & \textbf{0.42}               & 0.55                                 & \textbf{0.70}                        & 1.20                                 & \textbf{1.07}                        & 0.77                                 & \textbf{1.13}                        \\ \hline
            {\color[HTML]{2F75B5} noimu(w attention)} & {\color[HTML]{2F75B5} \textbf{0.69}} & {\color[HTML]{2F75B5} \textbf{0.97}} & {\color[HTML]{2F75B5} 0.68}          & {\color[HTML]{2F75B5} \textbf{1.04}} & {\color[HTML]{2F75B5} 0.73}          & {\color[HTML]{2F75B5} \textbf{0.66}} & {\color[HTML]{2F75B5} 0.64} & {\color[HTML]{2F75B5} \textbf{0.58}} & {\color[HTML]{2F75B5} \textbf{0.78}} & {\color[HTML]{2F75B5} \textbf{1.67}} & {\color[HTML]{2F75B5} \textbf{1.97}} & {\color[HTML]{2F75B5} \textbf{0.95}} & {\color[HTML]{2F75B5} \textbf{1.82}} \\
            noimu(w fc+activation)                    & 0.77                                 & 0.99                                 & \textbf{0.67}                        & 1.10                                 & \textbf{0.70}                        & 0.67                                 & \textbf{0.48}               & 0.80                                 & 0.80                                 & 2.36                                 & 2.07                                 & 1.04                                 & 2.21                                 \\ \hline
        \end{tabular}} 
    \end{center}
\end{table}
After extracting the features of the vertex map and the normal map, 
we add an additional self-attention module to improve the accuracy of pose estimation. 
The attention module can self-learn the importance of features, and give higher 
weight to more important features. We verify its effectiveness by comparing 
the result of the model which replaces the self-attention module with a single 
FC layer with activation function
(as formula (\ref{formula4})). The comparison is show in Table.~\ref{tab4}.
\begin{equation}
out=tanh(W_2(tanh(W_{1}x+b_1))+b_2).\label{formula4}
\end{equation}

\subsubsection{Loss function}
\begin{figure}[!t]
    \centering
    \includegraphics[width=6cm]{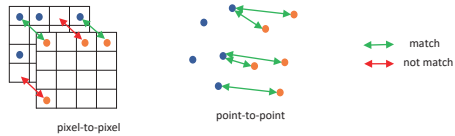}
    \caption{Matching points search strategy of Cho \emph{et al.}(pixel-to-pixel), our and Nubert \emph{et al.}(point-to-point).} 
    \label{fig6}
\end{figure}
\begin{table}[!t]
    \setlength{\belowcaptionskip}{-0.1cm}
    \caption{Comparison among different loss functions and matching point search strategy.}\label{tab5}
    \begin{center}
    \resizebox{1.0\textwidth}{!}{   
        \begin{tabular}{l|c|c|c|c|c|c|c|c|c|c|c|c|c}
            \hline
            $t_{rel}(\%)$                                                   & 00                                   & 01                                   & 02                                   & 03                                   & 04                                   & 05                                   & 06                                   & 07                                   & 08                                   & 09                                   & 10                                   & trainavg                              & testavg                              \\ \hline
            {\color[HTML]{2F75B5} imu(w   point-to-plane))+point-to-point}  & {\color[HTML]{2F75B5} \textbf{1.50}} & {\color[HTML]{2F75B5} \textbf{3.44}} & {\color[HTML]{2F75B5} \textbf{1.33}} & {\color[HTML]{2F75B5} x}             & {\color[HTML]{2F75B5} \textbf{0.94}} & {\color[HTML]{2F75B5} \textbf{0.98}} & {\color[HTML]{2F75B5} \textbf{0.90}} & {\color[HTML]{2F75B5} \textbf{1.00}} & {\color[HTML]{2F75B5} \textbf{1.63}} & {\color[HTML]{2F75B5} 2.24}          & {\color[HTML]{2F75B5} \textbf{1.83}} & {\color[HTML]{2F75B5} \textbf{1.58}} & {\color[HTML]{2F75B5} \textbf{2.03}} \\
            imu(w/o   point-to-plane)+point-to-point                        & 2.27                                 & 4.33                                 & 2.24                                 & x                                    & 1.59                                 & 1.70                                 & 1.26                                 & 1.29                                 & 1.87                                 & \textbf{2.04}                        & 2.07                                 & 2.07                                 & 2.06                                 \\
            imu(w   point-to-plane)+pixel-to-pixel                          & 2.14                                 & 4.36                                 & 2.29                                 & x                                    & 1.65                                 & 1.66                                 & 1.17                                 & 1.36                                 & 1.73                                 & 2.95                                 & 2.28                                 & 2.16                                 & 2.61                                 \\ \hline
            {\color[HTML]{2F75B5} noimu(w   point-to-plane)+point-to-point} & {\color[HTML]{2F75B5} \textbf{1.33}} & {\color[HTML]{2F75B5} \textbf{3.40}} & {\color[HTML]{2F75B5} \textbf{1.53}} & {\color[HTML]{2F75B5} \textbf{1.43}} & {\color[HTML]{2F75B5} 1.26}          & {\color[HTML]{2F75B5} 1.22}          & {\color[HTML]{2F75B5} 1.19}          & {\color[HTML]{2F75B5} \textbf{0.97}} & {\color[HTML]{2F75B5} 1.92}          & {\color[HTML]{2F75B5} 3.87}          & {\color[HTML]{2F75B5} \textbf{2.69}} & {\color[HTML]{2F75B5} \textbf{1.89}} & {\color[HTML]{2F75B5} \textbf{3.28}} \\
            noimu(w/o   point-to-plane)+point-to-point                      & 1.46                                 & 3.44                                 & 1.67                                 & 1.91                                 & \textbf{0.92}                        & \textbf{1.00}                        & \textbf{1.11}                        & 1.36                                 & \textbf{1.81}                        & 4.72                                 & 2.78                                 & 2.02                                 & 3.75                                 \\
            noimu(w   point-to-plane))+pixel-to-pixel                       & 2.76                                 & 4.43                                 & 2.73                                 & 2.07                                 & 1.71                                 & 1.50                                 & 1.32                                 & 1.32                                 & 1.95                                 & \textbf{3.68}                        & 3.65                                 & 2.47                                 & 3.67                                 \\ \hline \hline
            $r_{rel}(^\circ/100m)$                                          & 00                                   & 01                                   & 02                                   & 03                                   & 04                                   & 05                                   & 06                                   & 07                                   & 08                                   & 09                                   & 10                                   & trainavg                              & testavg                              \\ \hline
            {\color[HTML]{2F75B5} imu(w   point-to-plane))+point-to-point}  & {\color[HTML]{2F75B5} \textbf{0.70}} & {\color[HTML]{2F75B5} \textbf{0.99}} & {\color[HTML]{2F75B5} \textbf{0.59}} & {\color[HTML]{2F75B5} x}             & {\color[HTML]{2F75B5} \textbf{0.78}} & {\color[HTML]{2F75B5} \textbf{0.56}} & {\color[HTML]{2F75B5} \textbf{0.45}} & {\color[HTML]{2F75B5} \textbf{0.54}} & {\color[HTML]{2F75B5} \textbf{0.78}} & {\color[HTML]{2F75B5} \textbf{1.13}} & {\color[HTML]{2F75B5} \textbf{1.14}} & {\color[HTML]{2F75B5} \textbf{0.76}} & {\color[HTML]{2F75B5} \textbf{1.14}} \\
            imu(w/o   point-to-plane)+point-to-point                        & 1.01                                 & 1.12                                 & 0.98                                 & x                                    & 0.96                                 & 0.82                                 & 0.62                                 & 0.78                                 & 0.86                                 & 1.14                                 & 1.19                                 & 0.95                                 & 1.16                                 \\
            imu(w   point-to-plane)+pixel-to-pixel                          & 0.96                                 & 1.11                                 & 0.96                                 & x                                    & 0.98                                 & 0.83                                 & 0.58                                 & 0.83                                 & 0.87                                 & 1.52                                 & 1.27                                 & 0.99                                 & 1.39                                 \\ \hline
            {\color[HTML]{2F75B5} noimu(w   point-to-plane)+point-to-point} & {\color[HTML]{2F75B5} \textbf{0.69}} & {\color[HTML]{2F75B5} \textbf{0.97}} & {\color[HTML]{2F75B5} \textbf{0.68}} & {\color[HTML]{2F75B5} \textbf{1.04}} & {\color[HTML]{2F75B5} 0.73}          & {\color[HTML]{2F75B5} 0.66}          & {\color[HTML]{2F75B5} 0.64}          & {\color[HTML]{2F75B5} \textbf{0.58}} & {\color[HTML]{2F75B5} 0.78}          & {\color[HTML]{2F75B5} 1.67}          & {\color[HTML]{2F75B5} 1.97}          & {\color[HTML]{2F75B5} \textbf{0.95}} & {\color[HTML]{2F75B5} 1.82}          \\
            noimu(w/o   point-to-plane)+point-to-point                      & 0.73                                 & 0.99                                 & 0.70                                 & 1.34                                 & \textbf{0.69}                        & \textbf{0.58}                        & \textbf{0.49}                        & 0.85                                 & \textbf{0.76}                        & 1.85                                 & \textbf{1.84}                        & 0.98                                 & 1.85                                 \\
            noimu(w   point-to-plane))+pixel-to-pixel                       & 1.10                                 & 1.16                                 & 1.11                                 & 1.40                                 & 1.03                                 & 0.76                                 & 0.62                                 & 0.78                                 & 0.89                                 & \textbf{1.56}                        & 2.05                                 & 1.13                                 & \textbf{1.80}                        \\ \hline
        \end{tabular}} 
    \end{center}
\end{table}
Cho \emph{et al.} \cite{Cho} adopt the strategy of using the points with the same pixel in last and current vertex map 
as the matching points. Although the calculation speed is fast, the matching points found in this way are likely to be 
incorrect. Therefore, we and Nubert \emph{et al.} \cite{Nubert} imitate ICP algorithm, using the nearest neighbor as the 
matching point(see Fig.~\ref{fig6}). Although we use the same loss functions and the same matching point search strategy 
(nearest neighbor) as Nubert \emph{et al.} \cite{Nubert}, we search in the entire point clouds 
space, and maintain the number of points in search space not too large by removing 
most of the ground points and operating voxel grids downsample on point clouds.  
The number of points even is only 1/3 of the points sampled by the 2D projection which used in \cite{Nubert}.
Table.~\ref{tab5} shows the necessity of two loss parts and strategy of searching matching points in 
the entire point clouds.

\section{Conclusion}
In this paper, we proposed UnDeepLIO, an unsupervised learning-based 
odometry network. Different from other unsupervised lidar odometry methods, 
we additionally used IMU to assist odometry task. There have been already many IMU and 
lidar fusion algorithms in the traditional field for odometry, and it has become a 
trend to use the information of both at the same time. 
Moreover, we conduct extensive experiments on kitti dataset and 
experiments verify that our method is competitive with the most advanced methods. 
In ablation study, we validated the effectiveness of each component of our model. In the future, we will study how to incorporate mapping steps 
into our network framework and conduct online tests.

\bibliographystyle{splncs04}
\bibliography{reference}

\end{document}